\documentclass[acmtog]{acmart}

\AtBeginDocument{%
  \providecommand\BibTeX{{%
    \normalfont B\kern-0.5em{\scshape i\kern-0.25em b}\kern-0.8em\TeX}}}

\usepackage{enumitem}
\usepackage{xspace}
\usepackage{graphicx} 
\usepackage{amsmath}
\usepackage{booktabs}
\usepackage{xcolor}
\usepackage{multirow}
\usepackage{pifont}

\newcommand{\methodname}{CMD\xspace}
\newcommand{\modelname}{CondMV\xspace}

\copyrightyear{2025}
\acmYear{2025}
\setcopyright{cc}
\setcctype{by}
\acmConference[SIGGRAPH Conference Papers '25]{Special Interest Group on Computer Graphics and Interactive Techniques Conference Conference Papers }{August 10--14, 2025}{Vancouver, BC, Canada}
\acmBooktitle{Special Interest Group on Computer Graphics and Interactive Techniques Conference Conference Papers (SIGGRAPH Conference Papers '25), August 10--14, 2025, Vancouver, BC, Canada}
\acmDOI{10.1145/3721238.3730722}
\acmISBN{979-8-4007-1540-2/2025/08}
\definecolor{limegreen}{HTML}{badc58}
\definecolor{myyellow}{HTML}{f6e58d}

\begin{document}

\title{\methodname: Controllable Multiview Diffusion for 3D Editing and Progressive Generation} 

\renewcommand{\thefootnote}{\fnsymbol{footnote}}

\author{Peng Li}
\affiliation{%
  \institution{The Hong Kong University of Science and Technology}
  \city{Hong Kong}
  \country{China}
}
\author{Suizhi Ma}
\affiliation{%
  \institution{Johns Hopkins University}
  \city{Baltimore}
  \country{United States}
}
\author{Jialiang Chen}
\affiliation{%
  \institution{The Hong Kong University of Science and Technology}
  \city{Hong Kong}
  \country{China}
}
\author{Yuan Liu}
\authornotemark[2]
\affiliation{%
  \institution{The Hong Kong University of Science and Technology}
  \city{Hong Kong}
  \country{China}
}

\author{Congyi Zhang}
\affiliation{%
  \institution{University of British Columbia}
  \city{Vancouver}
  \country{Canada}
}
\author{Wei Xue}
\affiliation{%
  \institution{The Hong Kong University of Science and Technology}
  \city{Hong Kong}
  \country{China}
}
\author{Wenhan Luo}
\authornotemark[2]
\affiliation{%
  \institution{The Hong Kong University of Science and Technology}
  \city{Hong Kong}
  \country{China}
}
\author{Alla Sheffer}
\affiliation{%
  \institution{University of British Columbia}
  \city{Vancouver}
  \country{Canada}
}
\author{Wenping Wang}
\affiliation{%
  \institution{Texas A\&M University}
  \city{College Station}
  \country{United States}
}
\author{Yike Guo}
\affiliation{%
  \institution{The Hong Kong University of Science and Technology}
  \city{Hong Kong}
  \country{China}
}
\renewcommand{\shortauthors}{Li et al.}


\begin{abstract}
Recently, 3D generation methods have shown their powerful ability to automate 3D model creation. However, most 3D generation methods only rely on an input image or a text prompt to generate a 3D model, which lacks the control of each component of the generated 3D model. Any modifications of the input image lead to an entire regeneration of the 3D models. In this paper, we introduce a new method called \methodname that generates a 3D model from an input image while enabling flexible local editing of each component of the 3D model. In \methodname, we formulate the 3D generation as a conditional multiview diffusion model, which takes the existing or known parts as conditions and generates the edited or added components. This conditional multiview diffusion model not only allows the generation of 3D models part by part but also enables local editing of 3D models according to the local revision of the input image without changing other 3D parts. Extensive experiments are conducted to demonstrate that \methodname decomposes a complex 3D generation task into multiple components, improving the generation quality. Meanwhile, \methodname enables efficient and flexible local editing of a 3D model by just editing one rendered image. Project page: \href{https://penghtyx.github.io/CMD/}{https://penghtyx.github.io/CMD/}.
\end{abstract}


\begin{teaserfigure}
\centering
\includegraphics[width=0.96\textwidth]{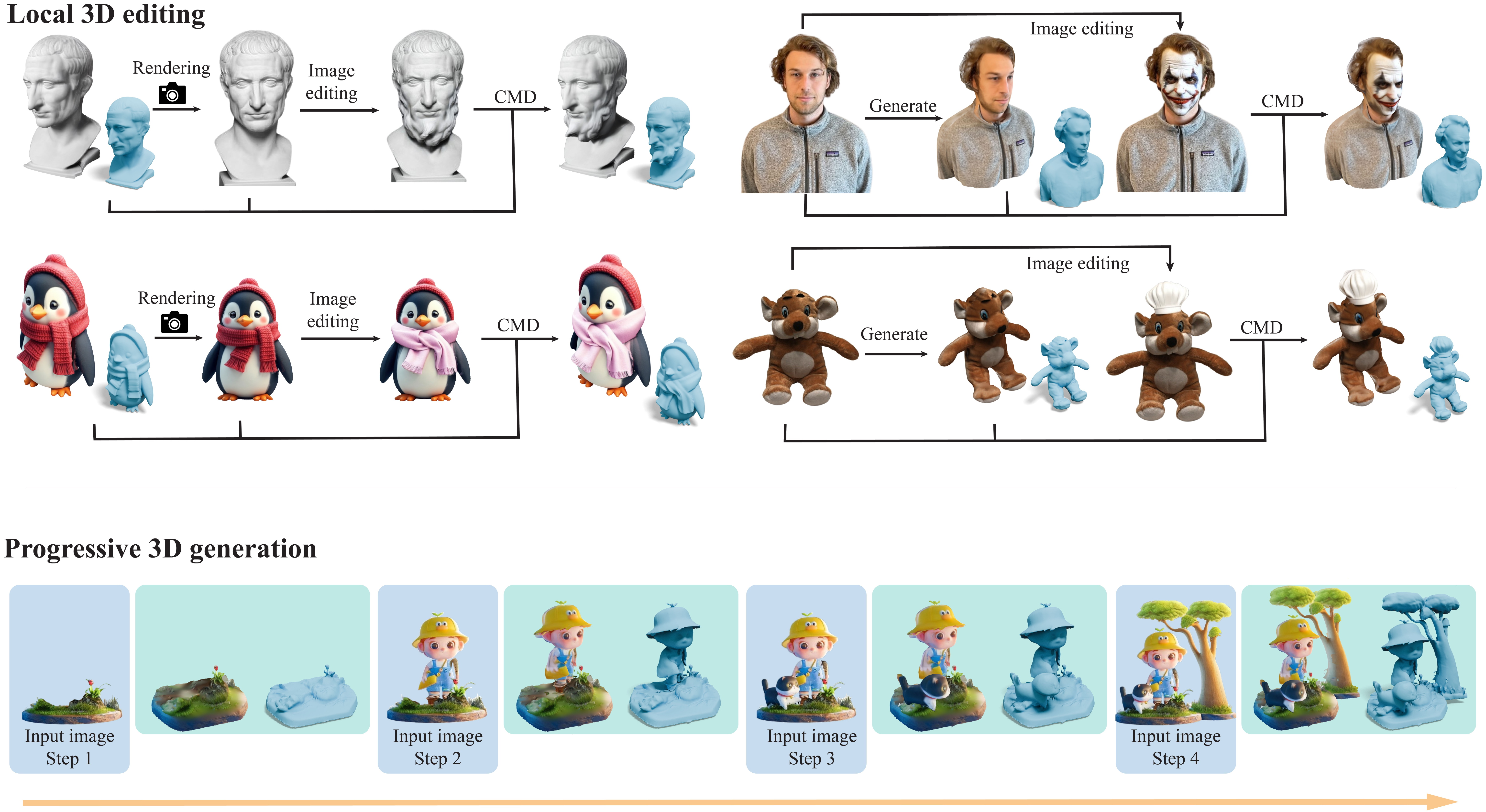}
  \caption{We present a novel conditional multiview diffusion model (\methodname) for (Top) easy-to-use local 3D editing of a 3D model by editing a rendered view and (Bottom) the single-view progressively generating a complex 3D model part by part with more fine details and structures. }
  \label{fig:teaser}
\end{teaserfigure}

\maketitle
\footnotetext[2]{Corresponding Author}
\section{Introduction}

\begin{table}
    \caption{Overview of the properties of 3D editing methods. 
    We consider the features of (a) image-based editing, (b) 3D-Guidance free, (c) high-quality mesh, (d) high-quality texture, and (e) running time. 
    Without any explicit 3D guidance, our method supports image-based 3D editing and outputs high-quality edited textured mesh, which strictly follows the given image reference. The whole process takes less than 20 seconds and exhibits significant efficiency against existing 3D editing approaches. }
    \begin{center}
        \begin{tabular}{ccccc|l@{\hskip 0cm}r}
        (a) & (b) & (c) & (d) & (e) &  \multicolumn{2}{c}{Methods} \\
        \hline
            \ding{55} &\ding{51} &\ding{51} &\ding{55} & 15min & TextDeformer & \small\textit{SIGG'23}  \\
            \ding{55} &\ding{51} &\ding{55} &\ding{51} & 46min & Vox-E & \small\textit{ICCV'23} \\
            \ding{55} &\ding{55} &\ding{51} &\ding{51} & 120min & MagicClay & \small\textit{SIGG Asia'24} \\
            \ding{51} &\ding{55} &\ding{55} &\ding{51} &  67min & TIP-editor & \small\textit{SIGG'24} \\
            \ding{55} &\ding{51} &\ding{55} &\ding{51} &  4min & DGE & \small\textit{ECCV'24} \\
            \ding{55} &\ding{55} &\ding{55} &\ding{51} &  - & Coin3D & \small\textit{SIGG'24} \\
        \hline
        \ding{51} &\ding{51} &\ding{51} &\ding{51} & 20s & \textbf{\methodname(ours)} \\
        \end{tabular}
    \end{center}
    
    \label{tab:features}
\end{table}

Recent advancements in 3D generation technologies~\cite{poole2022dreamfusion,liu2023syncdreamer,hong2023lrm,xu2024instantmesh,long2024wonder3d,zhang2024clay,li2024era3d,xiang2024structured} have shown remarkable potential for automating 3D model creation, enabling the generation of high-quality 3D models from text prompts or input images using diffusion generative models~\cite{ho2020denoising,rombach2022ldm} and neural representations~\cite{park2019deepsdf,wang2021neus,mildenhall2021nerf}. The achievement significantly advances downstream applications in areas like AR/VR, robotics, and manufacturing. Typically, a 3D generation pipeline involves generating multiview representations~\cite{liu2023syncdreamer,long2024wonder3d} from input text or images, followed by 3D shape generation~\cite{zhang2024clay, xiang2024structured} or reconstruction~\cite{mildenhall2021nerf,wang2021neus,palfinger2022continuous} to produce detailed 3D meshes.

While current 3D generation methods demonstrate impressive capabilities in producing high-quality 3D meshes, they lack flexibility when it comes to 3D editing. In a typical 3D modeling workflow, designers often need to iteratively refine 3D models for specific visual or functional requirements. 
This process demands the ability to make localized edits to the generated models. However, existing 3D generation frameworks~\cite{liu2023syncdreamer,long2024wonder3d,zhang2024clay,xiang2024structured} are primarily designed to create entire 3D models from 2D images and do not support localized modifications. Any minor changes to the input image require regenerating the entire 3D model, which not only risks altering unmodified regions but is also inefficient and unreliable for practical use.

Recent works~\cite{gao2023textdeformer,sella2023voxe,Barda24MagicClay,zhuang2024tip-editor,chen2024dge,dong2024coin3d} have attempted to address 3D editing challenges by introducing dedicated tools; however, they still fall short in terms of flexibility and efficiency. As summarized in Table~\ref{tab:features}, most existing 3D editing methods~\cite{sella2023voxe,Barda24MagicClay,chen2024dge,mvedit2024} rely on text prompts as inputs, using text-to-image diffusion models to modify selected regions of a 3D model based on the provided descriptions. While this approach facilitates basic edits, it falls short of providing the precision required to create specific appearances or shapes. Moreover, many methods~\cite{Barda24MagicClay,chen2024gaussianeditor,dong2024coin3d} require users to manually define allowed modification regions within the 3D space, posing additional challenges for novice. An another limitation is the inefficiency of these tools, as they usually utilize score distillation sampling(SDS)~\cite{poole2022dreamfusion} to distill a neural radiance or Gaussian~\cite{kerbl3Dgaussians} representation from the pre-trained 2D text-to-image models, which typically takes tens of minutes to edit even a small region, making iterative modifications impractical. These limitations in precision, usability, and efficiency render current 3D editing methods insufficient for meeting the demands of iterative and precise 3D model refinement. Even worse, many of these editing methods only support geometry or appearance modification but not both simultaneously.

In this paper, we address the above problems by introducing a new 3D generation method called \methodname that supports image-based 3D geometry and texture editing in $\sim$20 seconds. Our method is inspired by recent multiview generation models~\cite{liu2023syncdreamer,long2024wonder3d,li2024era3d}, which produces multiple color and normal images from a single color image. The core insight of \methodname is that the 3D model editing can be decomposed as multiview renderings editing and 2D-to-3D lifting. Therefore, given an existing textured mesh, we first edit a rendered view and synchronize the edited view to novel views, then propagate these edits to the given 3D model with incremental remeshing. To achieve multiview consistency before and after editing, we extend ControlNet~\cite{zhang2023adding} to a multiview ControlNet and incorporate it into a multiview diffusion model to produce target novel views, which follow both the edits and the renderings of the original mesh. Without explicitly specified 3D guidance, our method is more intuitive for users. Furthermore, it can yield high-quality textured mesh within 20 seconds, which is significantly efficient against existing methods, as featured in Table~\ref{tab:features}.

Beyond local editing capabilities, \methodname demonstrates significant potential in generating complex 3D assets with high fidelity. While existing approaches~\cite{tang2024lgm, liu2023syncdreamer, xu2024instantmesh, liu2023zero123} excel at generating simple objects, they often struggle with complex 3D asset modeling, primarily due to the scarcity of large-scale, generatable 3D datasets. \methodname mitigates this limitation benefiting from the progressive generation characteristic that leverages existing 3D models as conditioning signals. Specifically, we first decompose the input complex image into multiple simpler components with an off-the-shelf segmentation model. These components are then processed sequentially, with each generation step conditioned on the results of the previous iteration. This progressive approach allows the diffusion model to focus on generating detailed geometry for individual components while maintaining coherence with previously generated parts. To ensure spatial consistency across components and in the final model, we incorporate a global conditioning mechanism that enhances the model's ability to understand and maintain proper spatial relationships.

We conduct comprehensive evaluations of \methodname on two tasks, including local 3D editing and progressive 3D generation. As shown in Fig.~\ref{fig:teaser}, Fig.~\ref{fig:comp_appe} and Fig.~\ref{fig:comp_recon}, our method enables not only precise and realistic local editing but also demonstrates strong capability in complex 3D object generation. Both qualitative and quantitative evaluations demonstrate that \methodname significantly outperforms existing methods in terms of editing quality, computational efficiency and generation fidelity. Our contribution can be summarized as follows. 1) We propose a novel conditional multiview generation framework, \methodname, enabling both efficient 3D editing and high-quality 3D generation. 2) We demonstrate state-of-the-art performance in image-based 3D editing, achieving unprecedented flexibility and efficiency with editing in merely 20 seconds; 3)  We present an effective generation pipeline with a global condition scheme for progressive complex 3D asset generation that outperforms existing methods.

\section{Related Work}
\begin{figure*}
    \centering
    \includegraphics[width=0.8\textwidth]{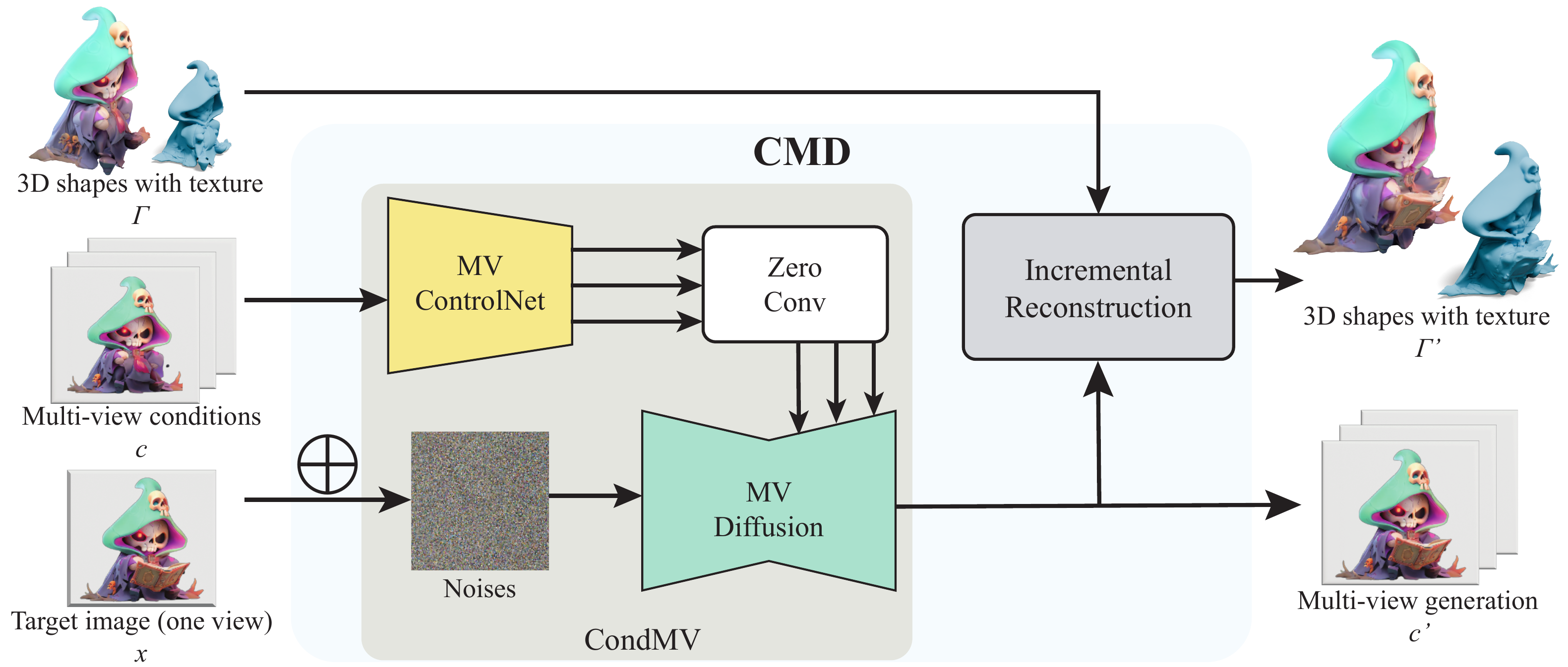}
    \caption{The overview of \methodname in local 3D editing. Our method takes a 3D mesh and an edited rendering (target image) of this mesh as input and produces the edited 3D meshes while keeping other regions unchanged. \methodname essentially consists of a \modelname that takes both target image and multiview conditions (RGB images and normal maps rendered from the given 3D mesh) as inputs and generates the multiview generations (RGB images and normal maps) that correspond to the target image. Then, \methodname incrementally reconstructs the output meshes from the multiview generations.
    }
    \label{fig:pipeline}
\end{figure*}
\paragraph{3D Generation.}

3D generation has witnessed significant progress in recent years, driven by the integration of neural radiance fields (NeRFs), implicit representations, and diffusion models. 2D-to-3D distillation methods leverage pre-trained 2D diffusion models to optimize 3D representations, avoiding the need for 3D training data. DreamFusion~\cite{poole2022dreamfusion} introduced Score Distillation Sampling (SDS) as a foundational approach, while later works like Magic3D~\cite{lin2023magic3d} improved generation quality through coarse-to-fine optimization, and ProlificDreamer~\cite{wang2024prolificdreamer} enhanced view consistency with variational score distillation. Fantasia3D~\cite{chen2023fantasia3d} improves geometry generation by decoupling geometry and appearance. However, these methods often require lengthy optimization and may produce artifacts due to imperfect 2D priors. To address this limitation, multiview generation approaches~\cite{shi2023zero123plus,liu2023syncdreamer,long2024wonder3d,li2024era3d,li2024pshuman,li2024mlrm} directly produce consistent multiview images from a single input, which are then reconstructed into 3D assets. These approaches bypass iterative SDS optimization, enabling faster generation, but their quality depends heavily on multiview consistency. Native 3D diffusion based methods ~\cite{zhang2024clay,wu2024direct3d,li2024craftsman,xiang2024structured} directly learn 3D representations, offering better geometric consistency. Despite significant advancements, these 3D generation methods remain predominantly object-centric and face considerable challenges when applied to complex scene generation.

\paragraph{SDS-based Mesh Editing.}

Score Distillation Sampling (SDS) enables text-driven 3D editing while mitigating the requirement for large scale 3D datasets. Related approaches adopt explicit~\cite{sella2023voxe} or implicit~\cite{zhuang2023dreameditor, Barda24MagicClay, rakotosaona2024nerfmeshing} neural representations to incorporate SDS loss for local editing. For fine-grained control, 2D diffusion models are integrated with localized strategies: InstructNeRF2NeRF~\cite{instructnerf2023} employs 2D diffusion to modify rendered NeRF views and subsequently updates the radiance field, whereas GaussianEditor~\cite{chen2024gaussianeditor} incorporates semantic segmentation for more precise edits. Other methods such as Progressive3D~\cite{cheng2023progressive3d}, FocalDreamer~\cite{li2023focaldreamer}, and NeRFInsert~\cite{sabat2024nerf} leverage localized SDS optimization to enable object insertion or region-specific refinement, while maintaining overall scene coherence. Additionally, proxy-guided techniques~\cite{dong2024coin3d, zhuang2024tip-editor, mvedit2024, mikaeili2023sked} offer intuitive user controls by leveraging coarse proxies, textual prompts, or sketches. Despite their impressive editing capabilities, these methods remain computationally inefficient.

\paragraph{Direct Mesh Editing.}
Traditional direct mesh editing methods focus on precision and interactivity through both commercial tools and geometric processing techniques. Popular software like ZBrush, Mudbox, and Substance Modeler provide intuitive interfaces for sculpting and refining meshes, enabling detailed and creative workflows. Research contributions include mesh deformation~\cite{skinningcourse:2014}, which supports smooth transformations, and local parametrization~\cite{schmidt2006interactive}, allowing for precise surface modification. Techniques like mesh simplification \cite{garland1997surface} and mesh subdivision \cite{catmull1998recursively} have further optimized mesh topology for applications in rendering and simulation. Example-based modeling has also significantly influenced this field. Methods like part assemblies \cite{funkhouser2004modeling} and statistical control of deformations~\cite{kalogerakis2012probabilistic} leverage large 3D databases for efficient shape composition. Recent methods have explored direct 3D editing from both structural and region-based perspectives: some approaches~\cite{bao2022neumesh, liu2023meshdiffusion} enable fine-grained mesh manipulation via vertex-level diffusion or disentangled mesh-guided latent representations, while others~\cite{chen2024dge, gao2024maskedlrm} rely on consistent multiview masks or priors from large-scale video models to guide region-specific editing. However, these methods only support either geometry or appearance editing. 

Our approach leverages the multi-view prior of off-the-shelf 3D generation models and introduces MVControlNet, a novel framework for efficient and flexible textured shape editing. Different from exiting MVControlNet based methods~\cite{li2025mvcontrol, oh2023controldreamer, chen2024textControlMesh, huang2024mvadapter, gu2025das}, which employ various geometry priors(like depth, normal, canny edge or camera ray) to generate pixel-wise aligned content, our method utilizes this structure to identify the edited regions automatically while preserving unedited areas, which are not simply pixel-wise aligned with the conditions as previous methods.
\section{Method}
\label{sec:method}
In this section, we present \methodname, a 3D generation method for efficient and precise local editing and high-quality 3D generation of complex 3D shapes. 
Our 3D generation method consists of two stages, a conditional multiview generation, denoted as \modelname(Section~\ref{subsec:multiview_generation}), that generates multiview images with given multiview constraints, and a differentiable rendering-based 3D reconstruction algorithm (Section~\ref{subsec:recon}) that lifts generated 2D multiview images to a 3D model.
We then elaborate on how to conduct two applications, 3D local editing (Sec.~\ref{subsec:editing}) and progressive 3D generation (Sec.~\ref{subsec:incremental}), with \methodname. For clarity, we take the 3D editing task as an example in the introduction of Sec.~\ref{subsec:multiview_generation} and Sec.~\ref{subsec:recon}. 

\subsection{Conditional Multiview Generation} 
\label{subsec:multiview_generation}
We decompose the 3D editing task into a multiview editing task using our conditional multiview diffusion model (\modelname).

\begin{figure}
    \centering
    \includegraphics[width=0.9\linewidth]{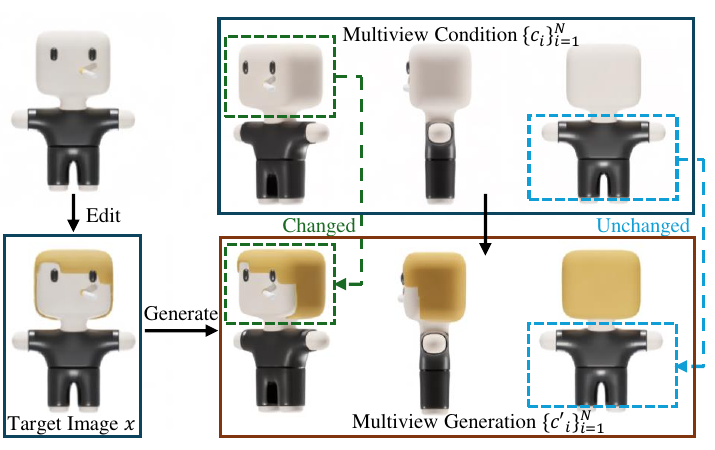}
    \caption{Input and output of \modelname. \modelname takes multiview conditions and target image as input and output multiview generation with only edited regions changed.}
    \label{fig:input-output}
\end{figure}

\paragraph{Input and Output}
As shown in Fig.~\ref{fig:input-output}, \modelname takes two inputs: a single-view target image $x$ and a set of multiview conditions $\{c_i\}_{i=1}^N$ consisting of color and normal maps $\{p_i\}_{i=1}^N$, $\{n_i\}_{i=1}^N$, rendered from a given 3D model $\Gamma$, where $N$ denotes the number of views and $c_i$ is the concatenation of $p_i$ and $n_i$. Specifically, the target image $x$ is a modified version of one rendered view from the original 3D model. 
Given these inputs, \methodname generates a set of color and normal maps $\{c'_i\}_{i=1}^N$ from six predefined viewpoints following \cite{li2024era3d,long2024wonder3d}. These generated maps are then used to reconstruct the edited 3D model $\Gamma'$. The generated multiview outputs $\{c'_i\}_{i=1}^N$ remain unchanged compared with the input conditions $\{c_i\}_{i=1}^N$ except in the edited area in $x$.

\paragraph{Cross-modality Multiview Diffusion}
\methodname is built upon a cross-modality multiview diffusion model that enables joint generation of multiple views for 3D generation. At its core, it incorporates a row-wise multiview attention mechanism between the self-attention and cross-attention layers of custom text-to-image (T2I) latent diffusion models (LDM) to enhance the cross-view consistency. Existing multiview diffusion models~\cite{liu2023syncdreamer, long2024wonder3d, wu2024unique3d} accept only a target image or text prompt as inputs, which fundamentally makes them unsuitable for editing tasks that require preserving specific regions of the given 3D model. Our \modelname leverages the multiview color and normal maps to represent the original 3D model, which thus enables fine-grained control over editing operations while maintaining global structural coherence.
In the following, we detail how to inject such a multiview condition in the cross-modality diffusion model.

\paragraph{Injecting Multiview Condition} ControlNet~\cite{zhang2023adding} is a controllable T2I diffusion model, which incorporates additional control signals~( e.g. edges, depth maps, or semantic maps) over the diffusion process to generate contents following the controls.  Drawing inspiration from it,
we propose a multiview ControlNet~(MVControlNet) to inject the conditions $c$ into the cross-modality diffusion model. Specifically, the structure and parameters of MVControlNet are copied from the pretrained backbone UNet of the base model~\cite{li2024era3d}. During training, we first employ a tiny convolutional neural network to transform the condition images $\{c_i\}_{i=1}^N$ to feature maps, which is then concatenated with $x$ and $\{\epsilon_i\}_{i=1}^N$ as the input of MVControlNet. Finally, each level of features from MVControlNet encoder is added to the corresponding decoder level of UNet through zero-convolution layers to obtain the edited cross-modality multiviews $\{c'_i\}_{i=1}^N$. The zero-convolution layers gradually learn to incorporate the multiview conditions while maintaining the backbone's original generation capabilities. Different from typical ControlNet, We finetune both the backbone denoising UNet and the MVControlNet simultaneously using the diffusion loss~\cite{ho2020denoising} to help our model identify the modified regions automatically while preserving unedited areas.

\subsection{Incremental Reconstruction}
\label{subsec:recon}

Given the edited multiview color and normal maps $\{c'_i\}_{i=1}^N$, we employ \textit{continuous remeshing}~\cite{palfinger2022continuous}, an efficient topology optimization approach that leverages differentiable rasterization~\cite{Laine2020diffrast} and Adam optimizer. While direct differentiable rendering with $\{n'_i\}_{i=1}^N$ supervision can lead to significant topology changes due to stochastic optimization, reconstructing the entire mesh from scratch is computationally inefficient, particularly when edits affect only a subset of vertices and faces. To mitigate these issues, we propose an incremental reconstruction strategy to reconstruct the edited models $\Gamma'$. Our approach initializes the optimization with the original mesh $\Gamma$ and employs differentiable rendering to minimize the following objective function
\begin{equation}
    \mathcal{L}_{recon} = \mathcal{L}_2(n'_i, \hat{n'}_i) + \mathcal{L}_2(\alpha_i, \hat{\alpha}_i) + \lambda \mathcal{L}_{\text{smooth}},
\end{equation}
where $n'_i$ and $\hat{n'}_i$ represent the generated normal maps and corresponding observations, respectively. We incorporate a mask alpha loss term that measures the difference between rasterized and generated foreground masks ($\hat{\alpha}_i$ and $\alpha_i$) to constrain the overall shape. Additionally, we apply Laplacian regularization $\mathcal{L}_{\text{smooth}}$ weighted by $\lambda$ to preserve mesh smoothness during optimization. The reconstruction process adaptively refines topology through iterative face splitting and merging operations to achieve optimal geometry. After geometry reconstruction, we bake the generated color maps $\{p'_i\}_{i=1}^N$ onto the mesh to obtain the textured 3D model $\Gamma'$. 

\begin{figure}[t]
    \centering\includegraphics[width=0.9\linewidth]{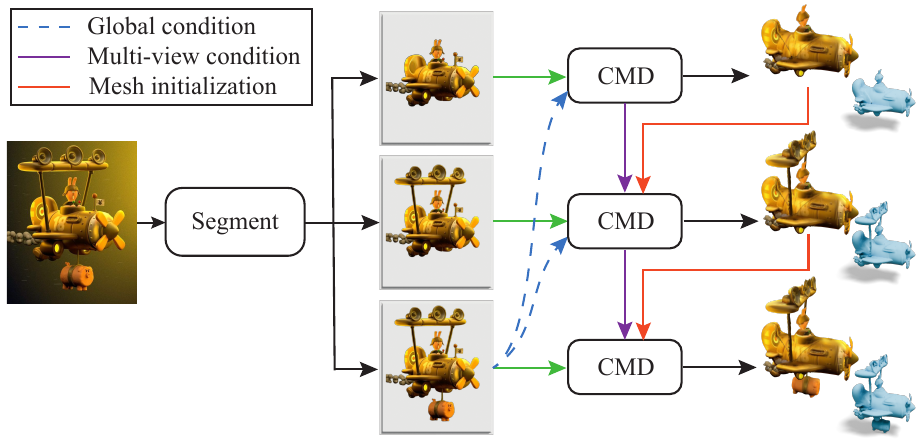}
    \caption{Progressive 3D generation pipeline. We decompose the input complex 3D shapes into several parts by image segmentation algorithm and then generate the shape in a part-by-part manner.}
    \label{fig:procedural_pipeline}
\end{figure}
\subsection{Application I: Local 3D Editing}
\label{subsec:editing}

Instead of manually specifying a set of allowed vertices or designing a suitable 3D shape proxy, \methodname enables efficient and realistic 3D textured model editing in a 3D-aware manner. Given a 3D model, we render both color and normal images from predefined six viewpoints with the azimuth of \{$0^\circ, 45^\circ, 90^\circ, 180^\circ, 270^\circ, 315^\circ$\} as the multiview conditions $\{c_i\}_{i=1}^N$. We then modify the $0^\circ$ view with an off-the-shelf image editing tool~\cite{openart} and take the edited result as target image $x$. The edited 3D model can be obtained through the \modelname followed by the incremental reconstruction. Note that our pipeline supports editing both existing and generated 3D models, as shown in Fig.~\ref{fig:teaser}.

\subsection{Application II: Progressive 3D Generation}
\label{subsec:incremental}

\begin{figure}
    \centering
    \includegraphics[width=0.8\linewidth]{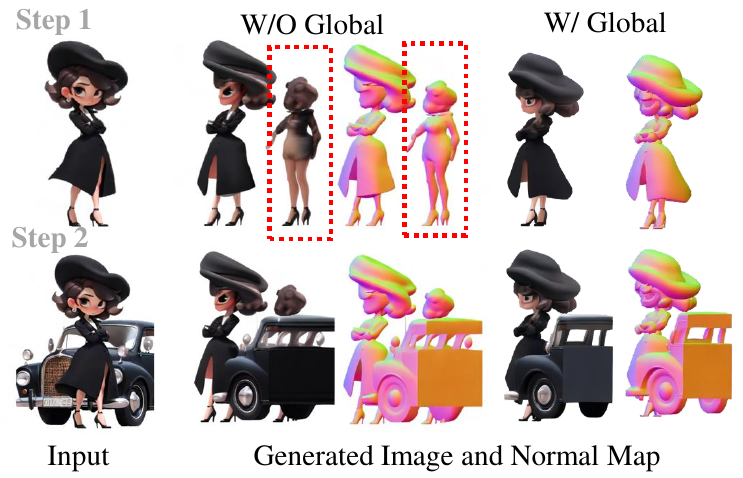}
    \caption {Effects of using global condition. ``W/O Global'' means not using the global condition. ``W/ Global'' means using the global condition. ``W/O Global'' leads to incorrect generation in Step 1 because the model is unaware of the car behind at this step (red bounding boxes). Using the full final target image leads to global aware generation at each step}.
    \label{fig:global_condition}
\end{figure}
\begin{figure*}
    \centering
    \includegraphics[width=\textwidth]{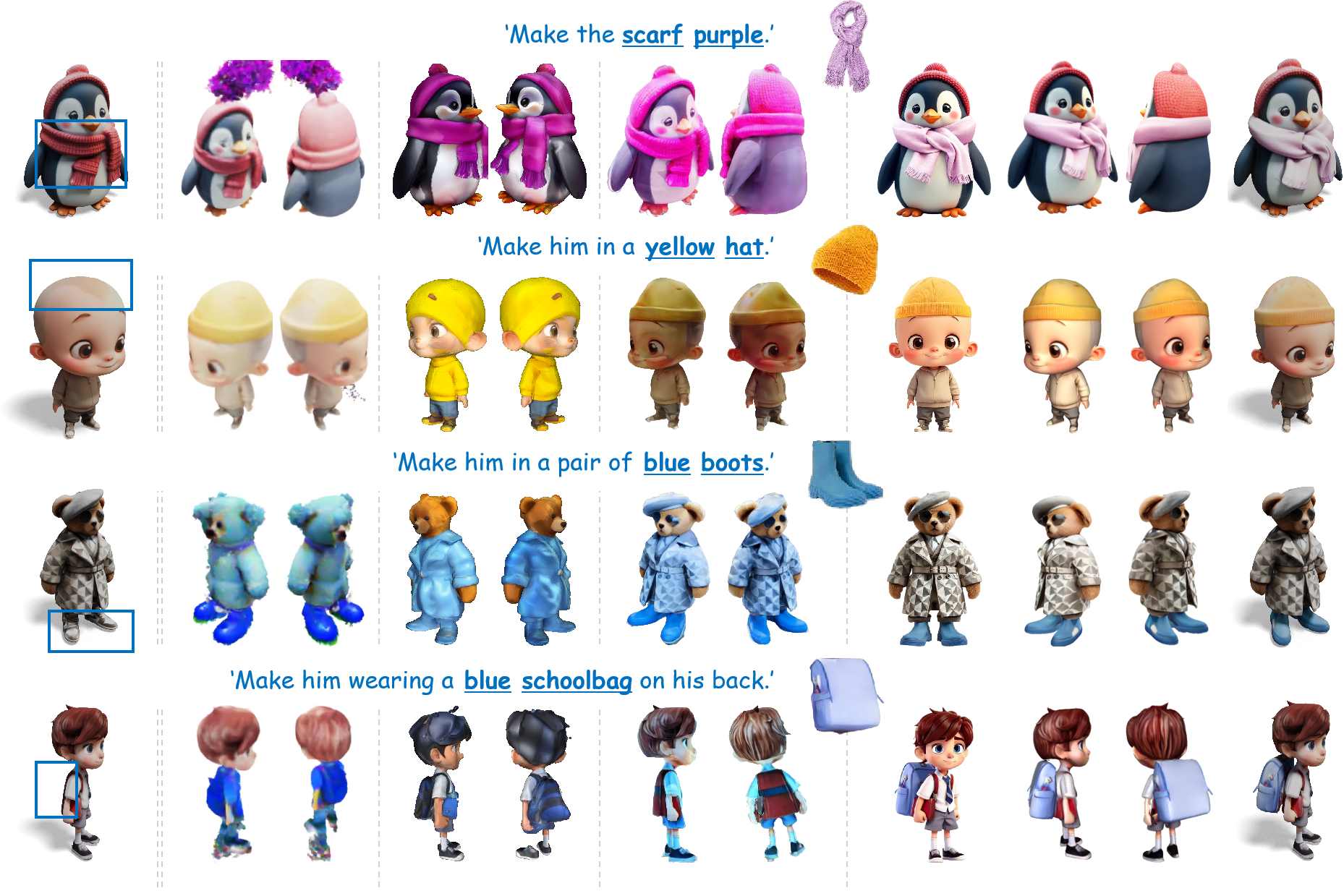}
    \leftline {\text {\small {\; Textured \, Mesh \qquad \quad \quad \; Vox-E \; \qquad \qquad \qquad \quad MVEdit \qquad \qquad \qquad \quad TIP-Editor \; \qquad \quad Edited View~(Input) \quad \qquad  Novel Views \quad \; \qquad Mesh}} }
    \caption{Qualitative comparisons of 3D appearance editing show that our method is capable of performing text- and/or image-based local editing while effectively preserving the uninstructed parts. The editing regions are highlighted with blue bounding boxes.}
    \label{fig:comp_appe}
\end{figure*}

Unlike most multiview-based generation models, which focus only on simple object generation, \methodname facilitates the progressive generation of complex 3D assets from single-view images in a progressive manner as shown in Fig.~\ref{fig:procedural_pipeline}. For this task, we use a segmentation model, e.g. SAM~\cite{kirillov2023segment}, to obtain the multi-step segmentation result and take them as step-by-step target image inputs $x$. In the first step, we set the condition $\{c_i\}_{i=1}^N$ as a set of white images and perform the reconstruction from scratch. In subsequent steps, we iteratively apply multiview generation and incremental reconstruction conditioned on results from the previous step. 

\paragraph{Global Condition.} 
Directly applying the above method for progressive generation leads to incorrect part sizes and layout. 
As illustrated in the middle of Fig.~\ref{fig:global_condition}, when we provide a partial component of the target image during initial generation, the diffusion model lacks spatial context regarding the component's intended position and size in the final reconstruction. This spatial ambiguity leads to positional and size incompatibilities in subsequent steps with conflicts, ultimately resulting in inconsistent multiple views.
To address this limitation, we introduce a global conditioning mechanism that incorporates the final target image to provide layout priors during each step generation. Specifically, we first employ VAE encoder of Stable Diffusion to transform the current step image and the global condition into their respective feature latents, which are then concatenated channel-wise with random noise and passed into the multiview diffusion model. To incorporate global condition, we expand the input convolution layer channels from 8 to 12, initializing the new layers with zeros for training.

\section{Experiments}
\label{sec:experiment}
\begin{figure}[t]
    \centering
    \includegraphics[width=0.5\textwidth]{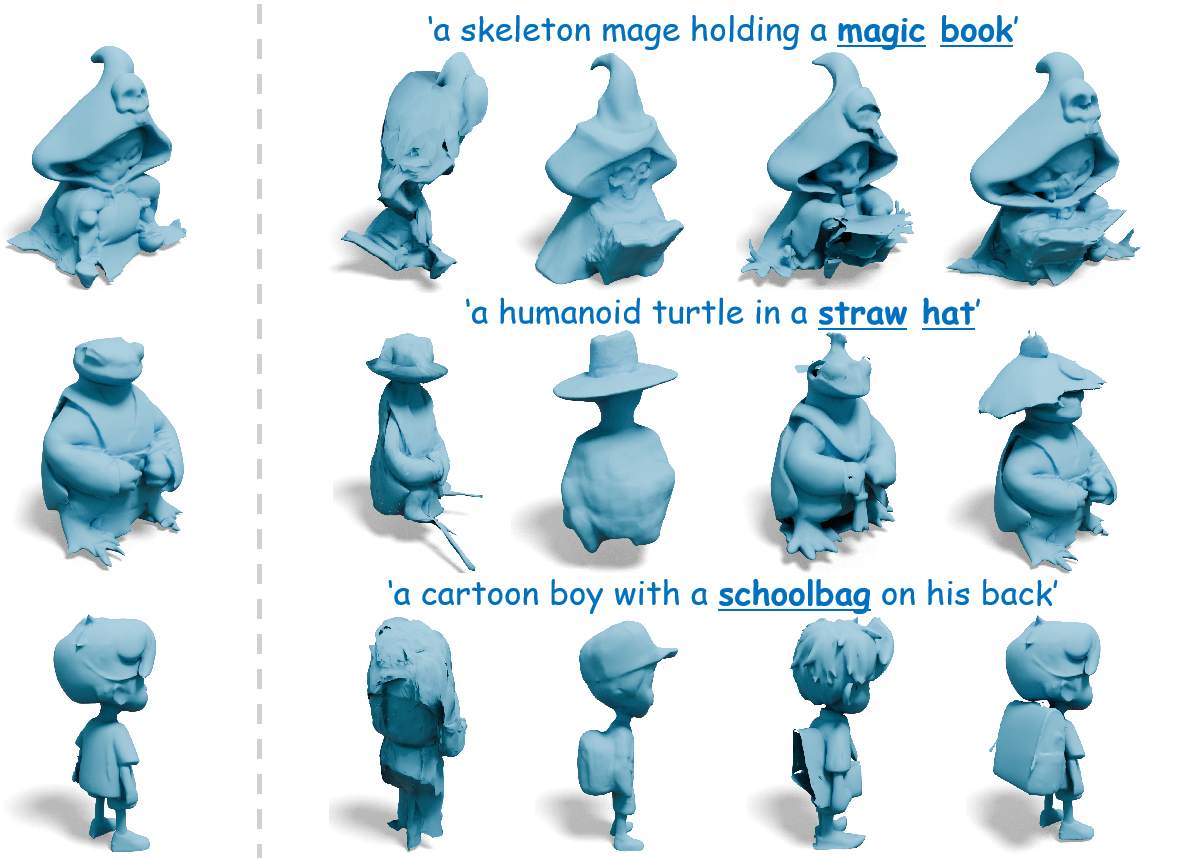}
    \leftline {\text {\small {\qquad Input \qquad \quad \; TextDeformer \quad MVEdit \; \quad MagicClay \, \qquad Ours}} }
    \caption{Qualitative comparisons of 3D geometry editing. }
    \label{fig:comp_mesh}
\end{figure}
\begin{table}
    \centering
    \caption{Quantitative comparison of 3D appearance editing methods. We evaluate text-image alignment using CLIP similarity scores (CLIP$_{sim}$). For a fair comparison, we augment TIP-Editor and DGE with an additional L1 constraint on the edited view (denoted as w/E) during the score distillation sampling process.}
    \setlength \tabcolsep{3pt}
    \begin{tabular}{c|cc|ccc}
        \toprule
        Method & TIP-Editor & DGE & TIP-Editor w/E & DGE w/E & Ours \\
        \midrule
        CLIP$_{sim}$ & 13.1 & 14.4 & 17.4 & 17.6 & \textbf{19.7}\\
        \bottomrule
    \end{tabular}
   
    \label{tab:app_clip}
\end{table}

\subsection{Experimental Setup}
Our training dataset is built upon the LVIS subset of Objaverse~\cite{deitke2023objaverse}, which comprises about $40,000$ 3D models. We augment these 3D models with part-level manipulation and object composition to train \methodname. Our evaluation dataset includes $20$ AI-generated 3D models (for editing task), and $30$ curated complex ones from the Internet (for generation task). We refer readers to the Appendix for more details about dataset setting and training details. 

\paragraph{Baselines.}
We compare \methodname against recent 3D editing approaches, including TextDeformer~\cite{gao2023textdeformer}, MVEdit~\cite{mvedit2024}, MagicClay~\cite{Barda24MagicClay}, which support geometry editing, as well as Vox-E~\cite{sella2023voxe}, TIP-Editor~\cite{zhuang2024tip-editor}, DGE~\cite{chen2024dge}, which are voxel- or radiance field-based and only output appearance. We also demonstrate the strength of the procedural generation pipeline by comparing with recent single image-based 3D generation methods, Wonder3D~\cite{long2024wonder3d}, InstantMesh~\cite{xu2024instantmesh}, Era3D~\cite{li2024era3d} and Unique3D~\cite{wu2024unique3d}. All baselines are evaluated using their official implementations with pretrained models.

\subsection{Local 3D Editing}
\begin{figure}
    \centering
    \includegraphics[width=\linewidth]{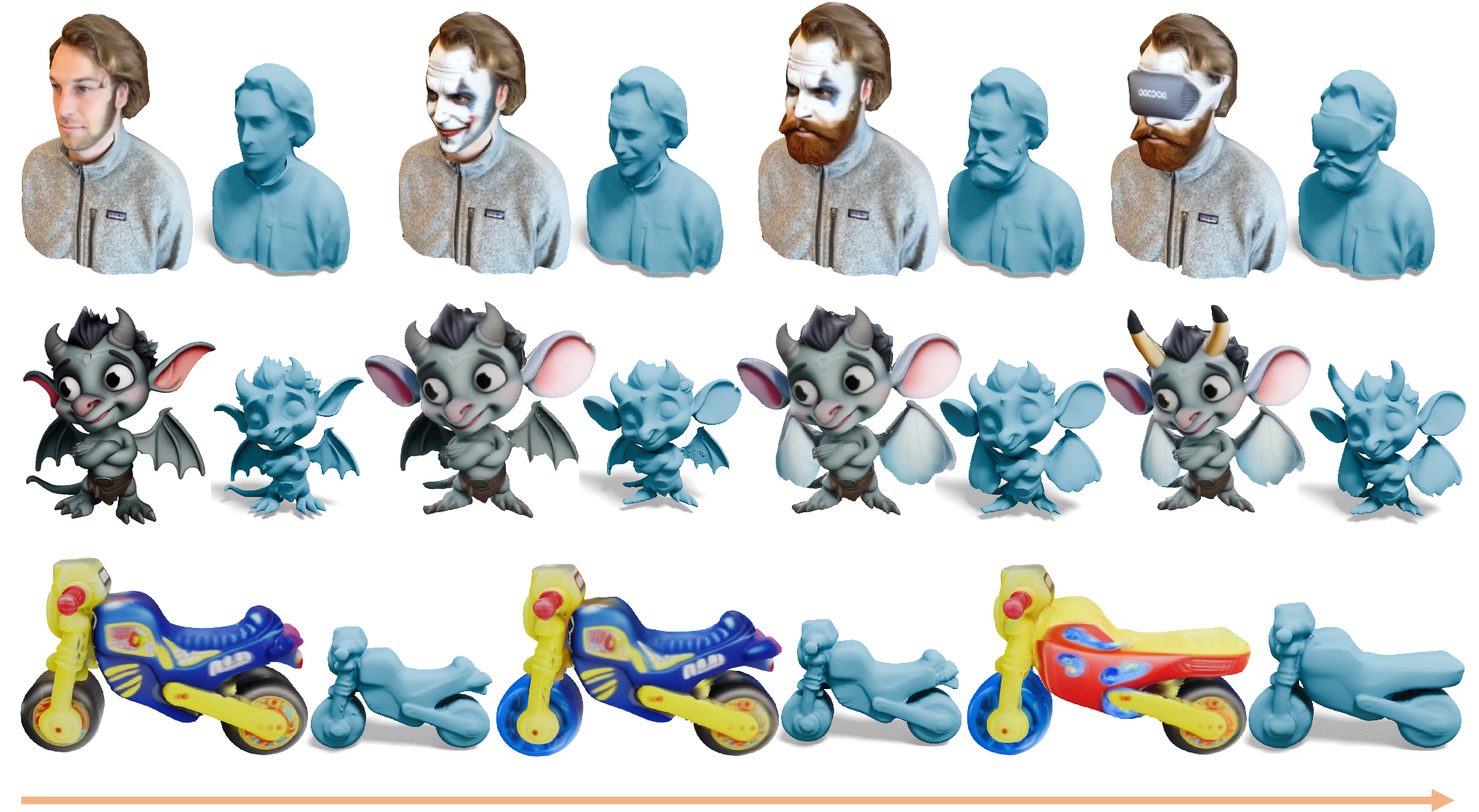}
    \caption{Sequential editing results. Our method facilitates sequential editing through the integration of multiple recursive local editing. At each stage, we are able to perform re-texturing, as well as local additions or modifications.}
    \label{fig:seque_edit}
\end{figure}

\begin{figure}
    \centering
    \includegraphics[width=\linewidth]{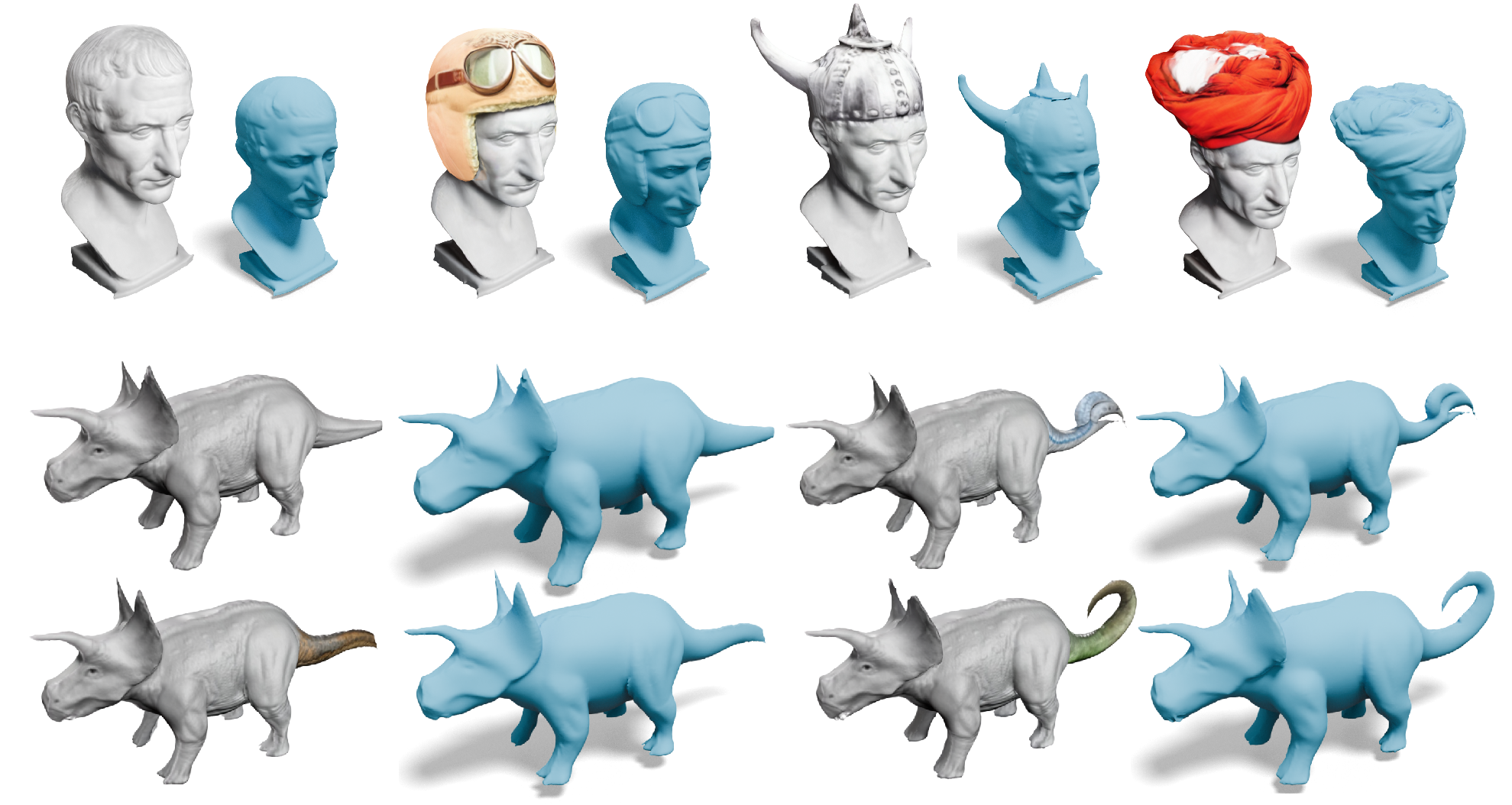}
    \caption{Diverse editing results. Our method supports diverse editing with different image prompts for given 3D models.}
    \label{fig:diverse_edit}
\end{figure}

\paragraph{Visual Results} In Fig.~\ref{fig:teaser}, Fig.~\ref{fig:seque_edit} and Fig.~\ref{fig:diverse_edit}, we present part of qualitative editing results of \methodname. Experiments on diverse 3D meshes demonstrate our capability in realistic and precise textured mesh manipulation, which successfully preserves geometric consistency with the input mesh while accurately reflecting the edited image. 

\paragraph{Geometry Comparisons.} We conduct comparative experiments against state-of-the-art methods (Fig.~\ref{fig:comp_mesh}). For a fair comparison with text-based baselines, we utilize ChatGPT to generate textual descriptions corresponding to our image prompts. Existing methods demonstrate limited controllability: TextDeformer and MVEdit fail to maintain geometric consistency in unedited regions, while MagicClay, even with manually specified editing regions, produces incomplete results with notable artifacts. In contrast, \methodname generates high-quality edits while maintaining strict geometric consistency with both input meshes and reference images.

\begin{figure*}
    \centering
    \includegraphics[width=0.95\textwidth]{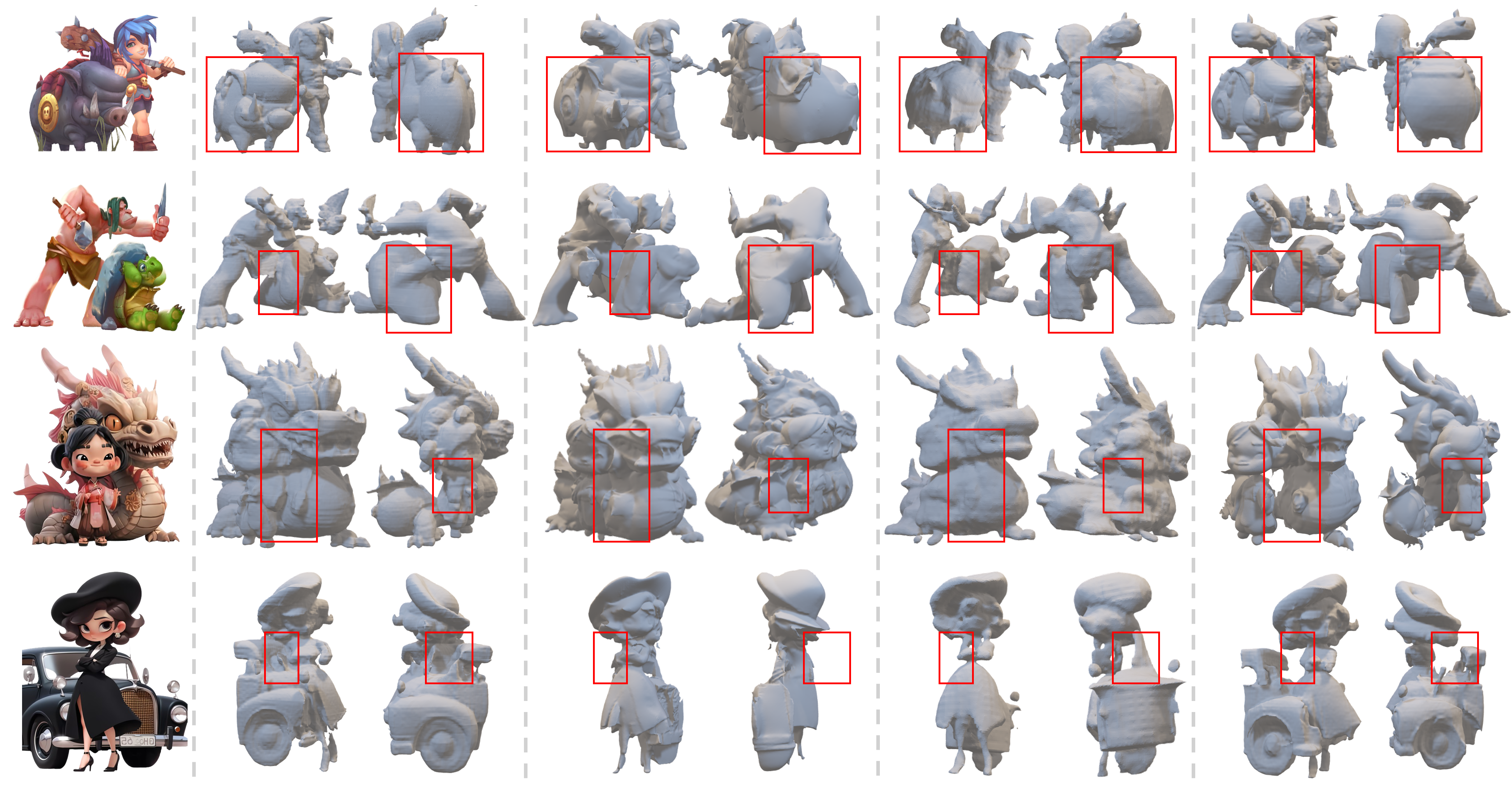}
    \leftline {\text {\small {\qquad \; Input \qquad \qquad \qquad \qquad \; \; Era3D \; \qquad \qquad \qquad  \qquad \qquad \; Unique3D \; \qquad \qquad \qquad \qquad \; InstantMesh \; \qquad \qquad \qquad \qquad \quad \; Ours}} }
    \caption{Qualitative comparisons of single-image 3D generation. Compared to baselines, our progressive generation pipeline demonstrates detailed local modeling and global coherence, showing the effectiveness of component division and enhanced detail carving in each component.}
    \label{fig:comp_recon}
\end{figure*}

\paragraph{Appearance Evaluation} Fig.~\ref{fig:comp_appe} presents appearance comparisons with both text- and image-guided approaches. Text-guided methods~(Vox-E and MVEdit) tend to generate entirely new objects rather than performing the desired local editing operation. For image-guided baselines~(TIP-Editor), we provide both image and text prompts. While TIP-Editor preserves object identity, it struggles with local editing, e.g., incorrectly applying global style transfer.

To quantitatively evaluate the alignment between the text prompt and edited results, we conduct experiments by randomly rendering eight views from $20$ edited textured meshes and computing the average CLIP similarity scores. For a fair comparison, we evaluate both the official implementations of baseline methods and enhanced versions incorporating our edited input view as additional constraints. Specifically, we augment the SDS process by introducing an L1 loss between the edited input view and the corresponding rendered view of the Gaussian splatting field. For TIP-Editor, this constraint is applied during its coarse editing stage. For DGE, it is integrated into the key frame editing process. These enhanced variants are denoted as TIP-Editor w/E and DGE w/E in Table~\ref{tab:app_clip}. The results demonstrate that \methodname achieves substantially higher text-image consistency compared to baseline methods while maintaining superior computational efficiency.

\paragraph{Editing Efficiency} A key advantage of \methodname lies in its computational efficiency. We benchmark our method against existing 3D editing approaches, including SDS-based and radiance field-based methods. Unlike prior works, our approach generates edited multi-view images in a single forward pass using 20-step DDIM denoising, followed by efficient incremental reconstruction, which directly outputs a mesh without requiring additional extraction.
The inference time of each component is detailed in Table~\ref{tab:inference_time}. Overall, \methodname achieves an 8-times speedup than state-of-the-art mesh editing methods~(Table~\ref{tab:features}), demonstrating its potential for interactive editing applications.  

\begin{table}
    \centering
    \caption{Inference time of our pipeline, including multi-view diffusion~(CondMV), incremental reconstruction and texture baking.}
    \setlength \tabcolsep{3pt}
    \begin{tabular}{c|cccc}
        \toprule
        Pipeline & CondMV & Reconstruction & Texture baking & Total \\
        \midrule
        Time/s & $\sim$9.1 & $\sim$9.3 & $\sim$1.6 & $\sim$20.0\\
        \bottomrule
    \end{tabular}
    \label{tab:inference_time}
\end{table}

\begin{table}[t]
  \centering
    \caption{Quantitative evaluation of Chamfer Distance, Volume IoU (for reconstruction), and LPIPS, SSIM, PSNR (for novel view synthesis). We compare our results with single image-based generation methods. "One-step generation" refers to employing CMD for generation directly without segmentation.}
  \setlength\tabcolsep{3pt}
  \begin{tabular}{l|cc|ccc}
    \toprule
    \multirow{2}{*}{Method} & \multicolumn{2}{c|}{Reconstruction} & \multicolumn{3}{c}{Novel View Synthesis} \\
    & CD ${\downarrow}$ & Vol. IoU ${\uparrow}$ & LPIPS ${\downarrow}$ & SSIM ${\uparrow}$ & PSNR ${\uparrow}$ \\ 
    \midrule
    Wonder3D        & 0.029  & 0.462  & 0.140  & 0.839  & 17.233 \\
    InstantMesh     & \underline{0.022}  & \underline{0.483}  & 0.131  & 0.846  & 17.476 \\
    Era3D           & 0.026  & 0.479  & 0.134  & 0.842  & 17.463 \\
    Unique3D        & 0.027  & 0.473  & \underline{0.127}  & \underline{0.853}  & \underline{17.512} \\
    \midrule
    Ours            & \textbf{0.017}  & \textbf{0.506}  & \textbf{0.121}  & \textbf{0.861}  & \textbf{17.681} \\
    w/o MVControlNet & 0.023  & 0.483  & 0.134  & 0.846  & 17.468 \\
    w/o Global Cond. & 0.019& 0.497  & 0.126  & 0.857  & 17.597 \\
    w/o Incre. Recon. & 0.017  & 0.501  & 0.124  & 0.857  & 17.673 \\
    One-step Generation & 0.023  & 0.486  & 0.134  & 0.848  & 17.473 \\
    \bottomrule
  \end{tabular}
  \label{tab:generation_comp}
\end{table}

\subsection{Progressive 3D Generation}
We provide quantitative comparison in Table~\ref{tab:generation_comp} and qualitative comparison in Fig.~\ref{fig:comp_recon}.
As reported in Table~\ref{tab:generation_comp}, \methodname significantly outperforms the baselines on all metrics, including the novel-view-synthesis and the geometry generation. The visual results in Fig.~\ref{fig:comp_recon} provide a more intuitive comparison. By decomposing a complex 3D generation task into several subtasks for each part, our method allows carving more details for each part while keeping all generated parts compatible to each other. In contrast, baseline methods either lose details in components or infer incorrect 3D layouts for different components. In the supplementary material, we also provide several examples of how our method supports an interactive progressive 3D mesh creation similar to Fig.~\ref{fig:teaser}. Our method progressively generates a complex 3D mesh part-by-part following users' 2D painting.

\begin{figure}
    \centering
    \includegraphics[width=\linewidth]{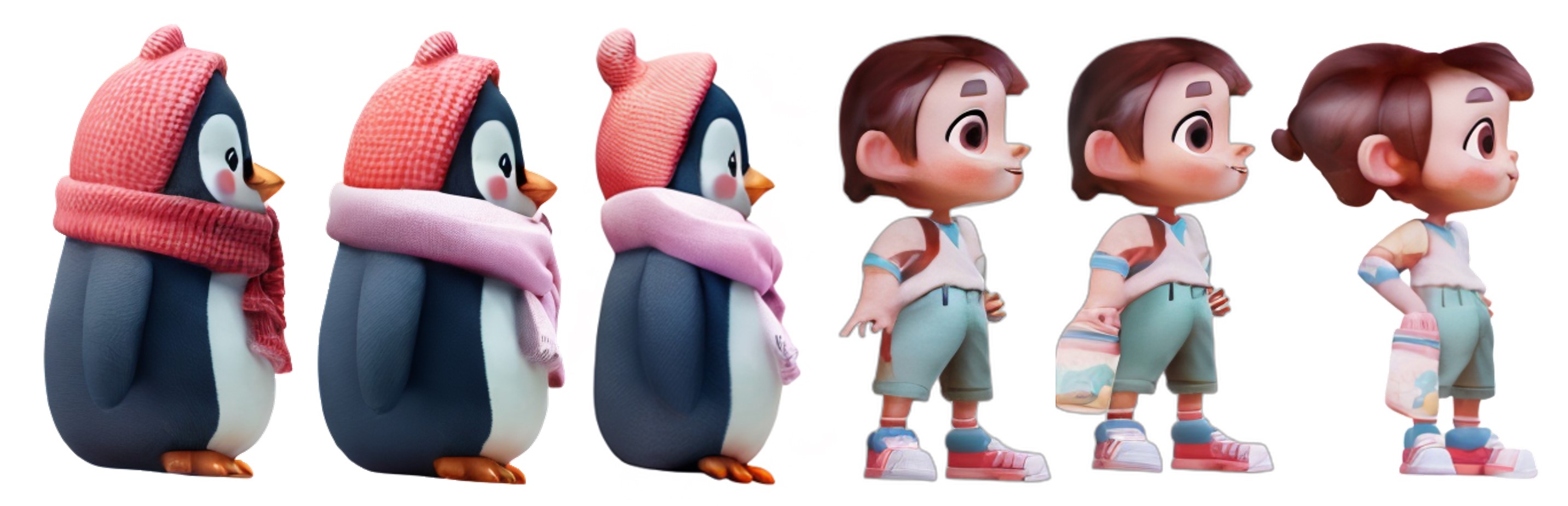}
    \small\leftline{\quad (a) Reference \qquad w/  \quad\qquad  w/o \quad (b) Reference \quad w/ \qquad\qquad w/o}
    \caption{Ablation study of MVControlNet in local 3D editing. In each case, we showcase the reference view of the original mesh, followed by the edited results w/ or w/o MVControlNet. CMD allows local editing while ensuring the consistency of other parts of the mesh.}
    \label{fig:abla_mvcontrol}
\end{figure}
\begin{figure}
    \centering
    \includegraphics[width=\linewidth]{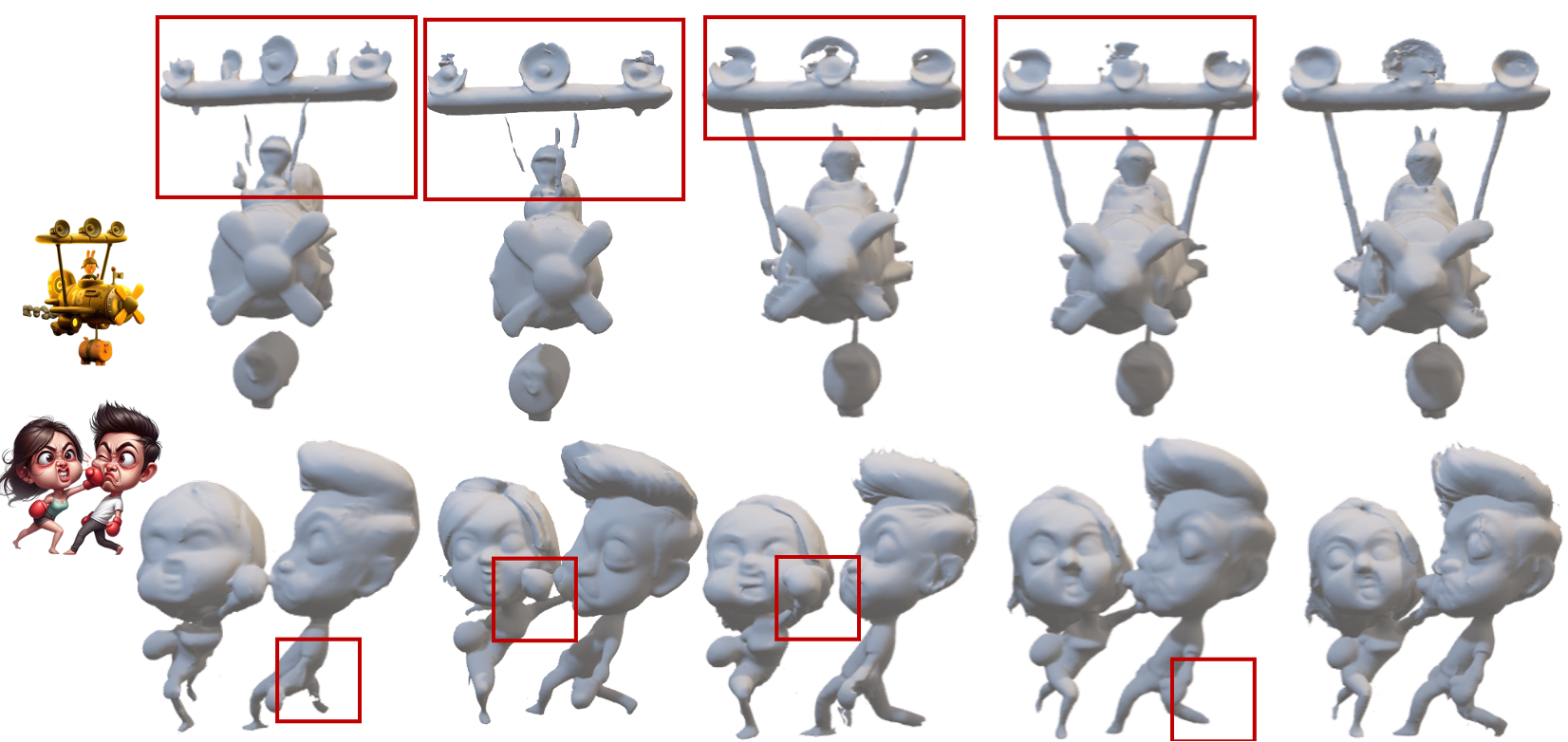}
    \footnotesize\leftline{\qquad\qquad Baseline \quad\quad w/o Segment.  + MVControlNet} \quad + Global Cond.\quad  Ours
    \caption{Ablation study of condition signals in progressive 3D generation. "Baseline" is the pretrained Era3D model. ‘w/o Segment.’ means using CMD for direct one-step generation without step-by-step segmentation. "Global Cond." means using the global image as an additional condition in the progressive 3D generation.}
    \label{fig:abla}
\end{figure}

\subsection{Ablation Studies}
\paragraph{MVControlNet for Local 3D Editing} We investigate the effect of our proposed MVControlNet in the local 3D editing task. Fig.~\ref{fig:abla_mvcontrol} compares the direct generation results using edited images and our controllable produced novel views. It is observed that our method shows strong controllability, allowing 3D-aware local editing while maintaining the 3D consistency of other parts of the original mesh, without requiring any explicit 3D guidance.

\paragraph{Control Signals in Progressive 3D Generation} In Fig.~\ref{fig:abla}, we perform a comprehensive ablation of our key designs for the progressive 3D generation. Starting from the Era3D~\cite{li2024era3d} baseline, we incrementally incorporate the key components. Adding MVControlNet enables step-wise generation, significantly improving local geometric details. However, this alone leads to potential inconsistencies between generation steps due to the lack of global context. Incorporating a global condition provides the crucial overall context for each generation step, effectively mitigating this issue and producing more coherent results. Our full pipeline with incremental reconstruction strategy further enhances fine-grain detail modeling.

\section{Limitations and Conclusions}

\paragraph{Limitations and Future Works.}
Despite the promising results, our method has several limitations. First, our pipeline relies on external image editing tools. The artifacts and unwanted modifications in imperfect image editing could lead to incorrect multiview generation. Another limitation is that our incremental reconstruction could not maintain the topology of the original mesh. It would be beneficial to automatically obtain the editing area and only update these faces. We leave this as a future work. 

\paragraph{Conclusion.}
In this paper, we present \methodname, a novel framework that enables both flexible local editing of 3D models and progressive generation of complex 3D assets. At the core of our method is a conditional multiview diffusion model that maintains global context while allowing precise local control. We further propose a global condition mechanism and incremental reconstruction strategy to enhance detail modeling. Through extensive experiments, we demonstrate that our approach significantly outperforms existing methods in terms of editing flexibility, generation quality, and computational efficiency. We believe our work represents an important step toward more practical and interactive 3D content creation.
\begin{acks}
The research was supported by
Theme-based Research Scheme (T45-205/21-N) from
Hong Kong RGC, Generative AI Research, Development Centre from InnoHK, in part by the National Natural Science Foundation of China (Grant No. 62372480), Guangdong Basic and Applied Basic Research Foundation (No.
2023A1515012839).

\end{acks}
\appendix
\section{Appendix}

\subsection{Datasets}
\label{sec:supp_dataset}
\paragraph{Training Dataset.} Our training dataset is built upon the LVIS subset of Objaverse~\cite{deitke2023objaverse}, which comprises about 40,000 3D models. For each model, we leverage Blender to render 8 pairs of color and normal images using orthogonal cameras, with azimuth angles uniformly distributed from 0$^{\circ}$ to 360$^{\circ}$ at a fixed elevation of 0$^{\circ}$. All the renderings have a resolution of 512$\times$ 512. To enable \methodname training, we augment the dataset with two strategies:
\begin{itemize}
    \item \textbf{Part-level Manipulation:} First, most of the 3D models in Objaverse are composed of multiple detachable parts. We sample 10,000 multi-component objects and randomly remove a part. We render paired multiview images for both the original and the modified models. 
    \item \textbf{Object Composition:} Second, We create 10,000 composite objects by randomly selecting and combining two to three objects from LVIS. Each object undergoes random scaling and translation transformations. We subsequently render paired multiviews for individual and composite objects with the same rendering setting.

\end{itemize}
The resulting dataset contains 60,000 3D models in total. During training, we employ a stratified sampling strategy with a ratio of 0.4:0.3:0.3 across the part-level dataset, composition ones, and original LVIS dataset, respectively, in which the LVIS dataset serves to maintain the training distribution of the base model. 

These datasets are curated to enhance the diffusion model with the ability to identify local modification and occlusions among multiple parts of complex objects. Therefore, it is unnecessary to ensure global semantically plausible. The base model itself could generate reasonable and plausible multiviews.

\begin{figure}[h]
    \centering
    \includegraphics[width=\linewidth]{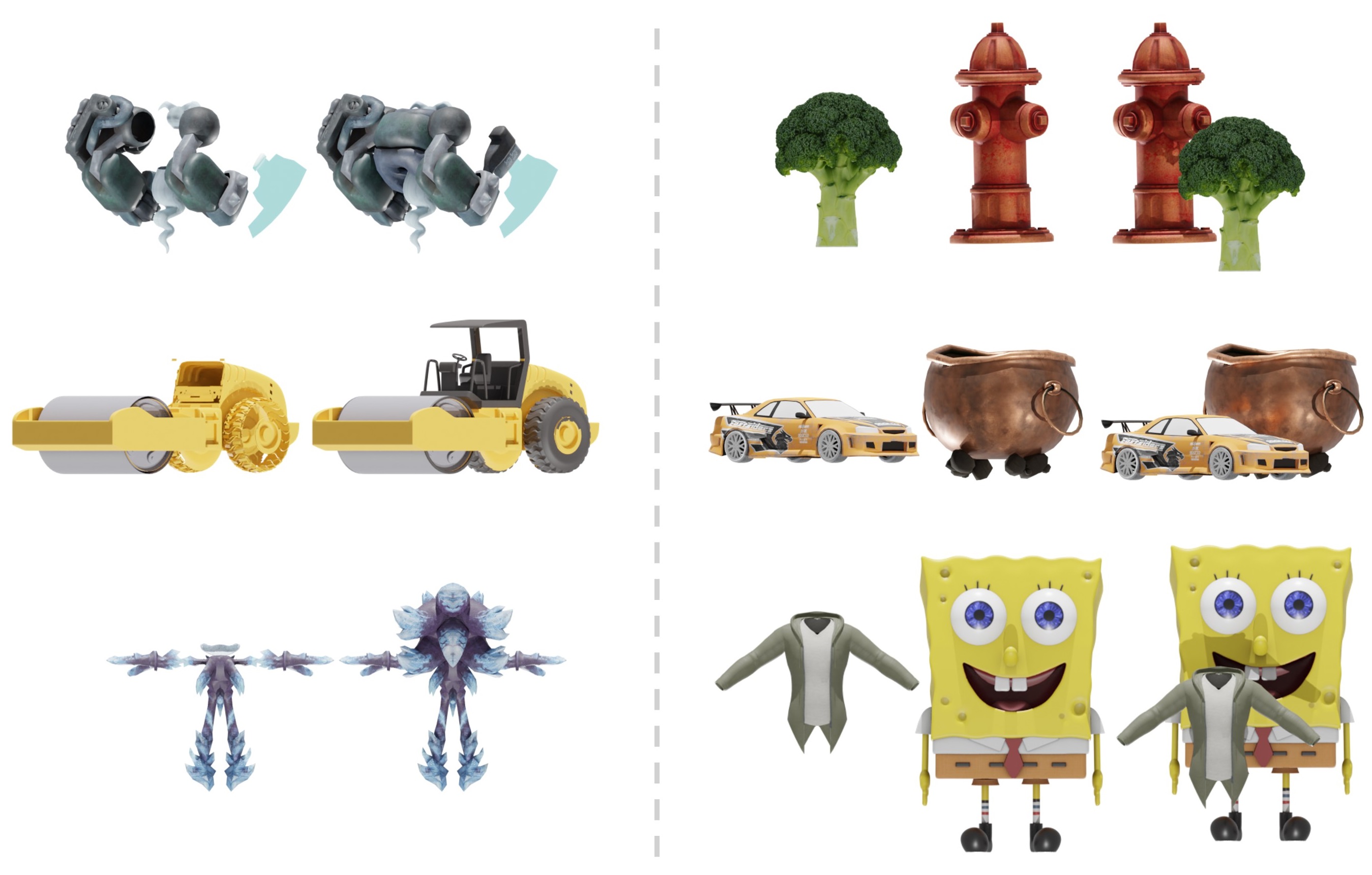}
    \small\leftline{\qquad (a) part-level manipulation \qquad\quad (b) multiple objects composition}
    \caption{Training dataset samples.}
    \label{fig:training_sample}
\end{figure}

\begin{figure}[h]
    \centering
    \includegraphics[width=\linewidth]{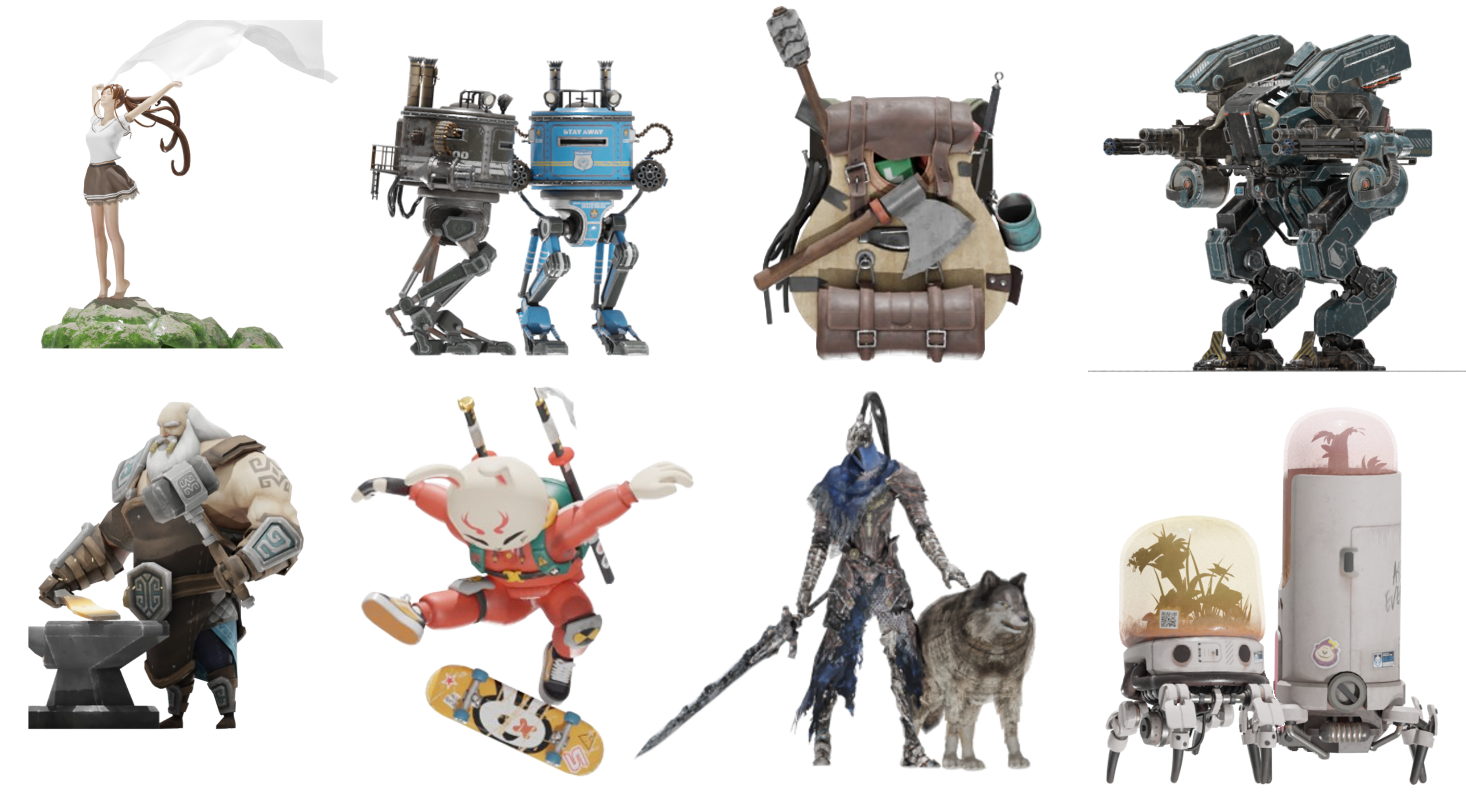}
    \caption{Testset samples.}
    \label{fig:testset_sample}
\end{figure}

\paragraph{Evaluation Dataset.} 
For the editing task, we curate a dataset comprising 15 object-centric images sourced from the web and 5 cases from widely used 3D benchmarks (Instruct-NeRF2NeRF~\cite{instructnerf2023} and DTU~\cite{jensen2014dtu}). We first process these images using GPT-4o to generate descriptive prompts, then reconstruct them as textured meshes using Era3D. The resulting meshes were rendered to serve as editable inputs and multiview conditioning data for CMD.

For the evaluation of \methodname's capabilities in the 3D generation of complex shapes, we curate a testset of 30 high-quality textured models from the website, which feature intricate multi-component structures that pose significant challenges for existing approaches. We show some examples in Fig.~\ref{fig:testset_sample}. These models are rendered from randomized viewpoints. 

\paragraph{Segmentation workflow.}
To facilitate \methodname inference in generation task, we develop a semantic-aware segmentation workflow to obtain step-by-step segmentation masks. Our segmentation workflow is built upon the Segment Anything Model~(SAM) \cite{kirillov2023segment}. Specifically, we first use SAM to generate several masks which are processed to be non-overlapping and collectively exhausted by coloring the masks in descending order of their areas. We then apply pretrained CLIP~\cite{radford2021learning} to obtain semantic embeddings for all patches which are dimensionally reduced by Principal Component Analysis (PCA) and concatenated with the center points and bounding boxes as patch features. Finally, we leverage KMeans to cluster the patches by features and convert the results to step-by-step segmentation masks. While this approach works well in most cases, optionally repeating the process and manually selecting the best one is also efficient. Since these are preprocessing steps to prepare inputs for our method and are not part of the method itself, we remain open to any alternatives for completing this task. After the segmentation, we simply follow the left-to-right and bottom-to-top order for progressive generation.

\subsection{Implementation Details.}

\paragraph{Training Details} Our implementation is built upon the open-source multiview diffusion model, Era3D~\cite{li2024era3d}, which is tuned from Stable Diffusion~(SD2.1-Unlip). We train \methodname on 4 H800 GPUs (80GB VRAM) using a batch size of 32 for 30,000 steps. The learning rate is set to 1e-4 and the training process takes approximately 40 hours. During inference, text-guided and image guidance are set to 3.0 and we leverage 20 sampling steps with DDIM. Following Unique3D~\cite{wu2024unique3d}, we adopt a coarse-to-fine strategy for incremental reconstruction, with each stage performing 100 steps of differentiable rendering for incremental reconstruction.

\paragraph{Multiple conditions drop strategy}

\begin{figure}[t]
    \centering
    \includegraphics[width=0.3\textwidth]{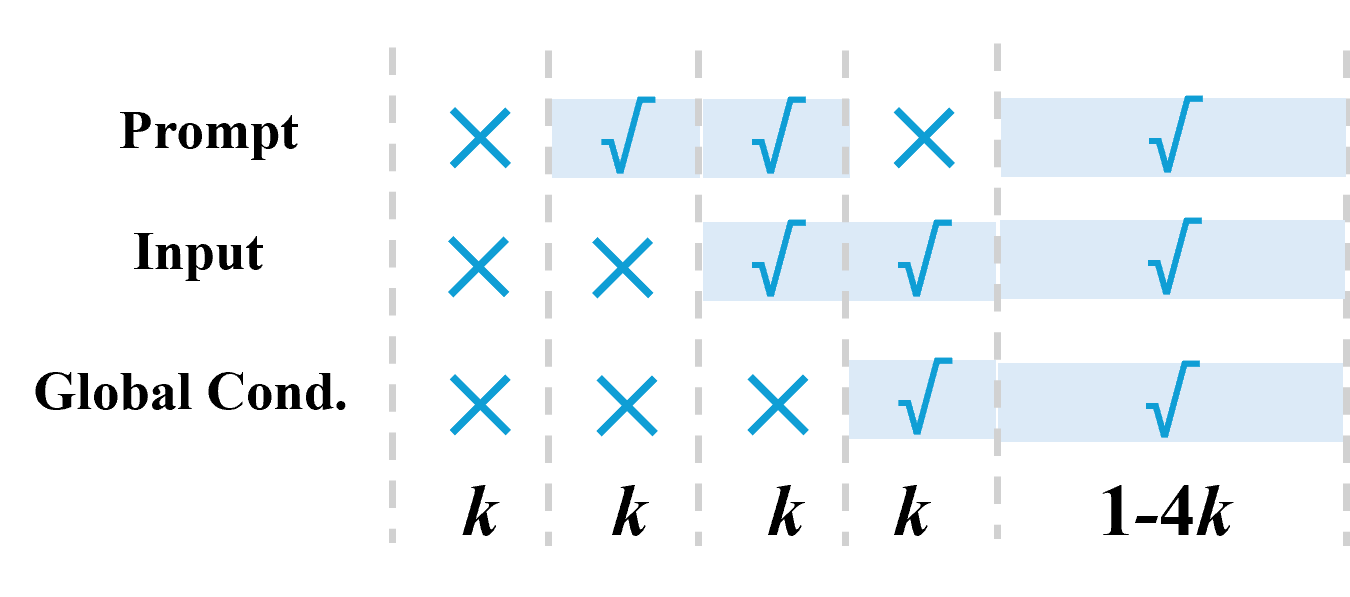}
    \caption{Dropping strategy for multiple conditions classifier-free guidance.}
    \label{fig:drop_strategy}
\end{figure}

Following Era3D, we condition the diffusion model on the view and domain~(color or normal) prompt for general guidance. Specifically, we feed the unified prompts below into the Unet of diffusion models via cross-attention:
\textit{"a rendering image of 3D models, {view}, {domain} map"}, where \textit{view} is selected from \{front, front right, right, back, left, front left\} and \textit{domain} is color or normal. 

Due to the introduction of multiple conditions, input image $x_k$, global condition $x$, and the prompt mentioned above, dropping them directly will weaken the influence of each component. To address this issue, we employ a mix-dropping strategy for multiple conditions classifier-free guidance training. As illustrated in Fig.~\ref{fig:drop_strategy}, we randomly drop the combination of three conditions during training. Considering the global condition is designed for the input image, we do not include the case of dropping the input image while keeping global condition. We empirically set k as 0.05 during training. In the editing task, there are no global condition images. To unify the framework, we set the global condition images the same as the target image $x$ in the editing task.

\paragraph{Incremental reconstruction for generation} To enhance the modeling capability for complex assets, during the incremental reconstruction step, we directly locate the approximate region of the newly added part by comparing the multiview masks between the current and previous step. We then initialize a sphere in this region and only update the topology of the sphere. This simple strategy avoids the artifacts near multiple parts connections caused by global differentiable rendering while maintaining the topology of previous reconstructions.

\paragraph{Evaluation Metrics.} Due to the lack of mesh editing ground truth, we only provide qualitative comparisons about geometry manipulation. For appearance editing, we follow Instruct-N2N~\cite{instructnerf2023} and use clip score to evaluate the alignment between the prompt and edited renderings. We also provide qualitative and quantitative comparisons of the generation task, including commonly used metrics, such as PSNR, SSIM, LPIPS, Chamfer Distance~(CD), and Volume IoU~(Vol. IoU).

\subsection{More Experiments}
\begin{table}
    \centering
    \setlength \tabcolsep{3pt}
    \begin{tabular}{c|cc|ccc}
        \toprule
        Method & MVEdit & TIP-Editor & DGE  & Ours \\
        \midrule
        Preference/ $\%$ & 12.7 & 8.7 & 30.2 &  \textbf{48.4}\\
        \bottomrule
    \end{tabular}
    \caption{User study on 3D editing methods. Our approach significantly outperforms other baselines in terms of human performance.}
    \label{tab:user_study}
\end{table}

\paragraph{User study.} To validate the superiority of our method in 3D editing, we conduct a comprehensive user study on an extended dataset of 40 examples, following the same processing protocol in the same manner described in Section~\ref{sec:supp_dataset}. For each case, we provide side-by-side rotation videos of edited results and corresponding prompt instructions. We invite 20 volunteers to choose the highest-quality output aligned with the prompts. As demonstrated in Table~\ref{tab:user_study}, the collected human feedback consistently favors our CMD approach over competing baselines. 

\subsection{Discussions}

\begin{figure}
    \centering
    \includegraphics[width=\linewidth]{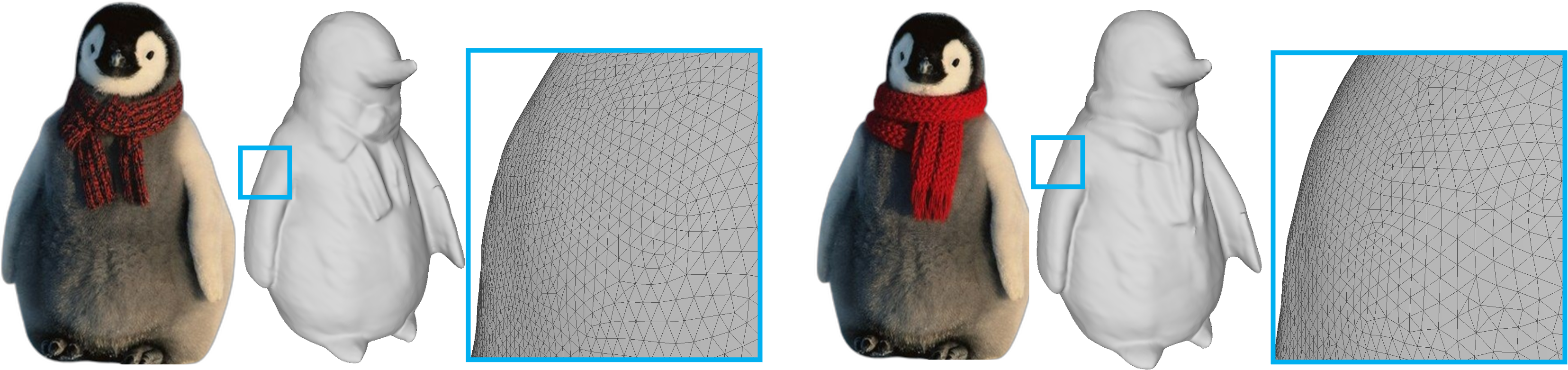}
    \leftline {\text {\small {\qquad\qquad\quad Before editing \qquad\qquad\qquad\qquad After editing}}}
    \caption{One limitation of CMD is that its incremental reconstruction fails to preserve the original mesh topology due to random optimization.}
    \label{fig:limitation_topo_change}
\end{figure}

\paragraph{Single view editing ambiguity.} Single-image editing faces inherent ambiguity. In practice application, our method first generates an initial edited result. If the user finds the novel-view edits unsatisfactory, they can interactively rotate the model and iteratively repeat the editing pipeline until the desired results are achieved. Therefore, our approach could mitigate ambiguity to some extent.

\paragraph{Mesh preservation after editing.} Our method preserve the identity of the original mesh in two ways. First, our MVControlnet allows generating consistent multiview normal and color maps before and after editing with only edited regions changed. Second, we incrementally reconstruct the geometry with initialization of the pre-step shape, which helps retain the details to a great extent. However, this pipeline could not maintain the fine-grain topology of the original mesh due to inherent stochasticity in the optimization process and continuous remeshing operations, as demonstrated in Figure~\ref{fig:limitation_topo_change}. A potential direction for improvement would involve developing automated techniques to localize editable regions and selectively update the corresponding mesh faces.

\bibliographystyle{ACM-Reference-Format}
\bibliography{ref}

\end{document}